\documentclass[11pt,a4paper]{article}
\usepackage[hyperref]{acl2019}
\usepackage{times}
\usepackage{latexsym}
\usepackage{url}
\usepackage{graphicx}
\usepackage{amsmath}
\usepackage{enumerate}
\usepackage{amsfonts}
\usepackage{array}
\usepackage{subfigure}
\usepackage{amssymb}
\usepackage{multirow}
\usepackage{booktabs}
\usepackage{footmisc}
\usepackage{floatrow}
\usepackage{mdwlist}
\usepackage{xcolor}
\usepackage{paralist}
\usepackage{diagbox}
\usepackage[linesnumbered,ruled,vlined]{algorithm2e}

\newcommand{\sts}{{{\textsc{Seq2Seq}}}\xspace}
\newcommand{\tabincell}[2]{\begin{tabular}{@{}#1@{}}#2\end{tabular}}

\aclfinalcopy 
\def\aclpaperid{1466} 

\title{Self-Supervised Dialogue Learning}
\author{Jiawei Wu \and Xin Wang \and William Yang Wang \\
Department of Computer Science\\
University of California, Santa Barbara\\
Santa Barbara, CA 93106 USA\\
{\tt \{jiawei\_wu,xwang,william\}@cs.ucsb.edu}}

\date{}

\begin{document}
\maketitle
\begin{abstract}
The sequential order of utterances is often meaningful in coherent dialogues, and the order changes of utterances could lead to low-quality and incoherent conversations.
We consider the order information as a crucial supervised signal for dialogue learning, which, however, has been neglected by many previous dialogue systems. 
Therefore, in this paper, we introduce a self-supervised learning task, \textit{inconsistent order detection}, to explicitly capture the flow of conversation in dialogues. Given a sampled utterance pair triple, the task is to predict whether it is ordered or misordered. 
Then we propose a sampling-based self-supervised network $\mathcal{SSN}$ to perform the prediction with sampled triple references from previous dialogue history. Furthermore, we design a joint learning framework where $\mathcal{SSN}$ can guide the dialogue systems towards more coherent and relevant dialogue learning through adversarial training. We demonstrate that the proposed methods can be applied to both open-domain and task-oriented dialogue scenarios, and achieve the new state-of-the-art performance on the OpenSubtitiles and Movie-Ticket Booking datasets.
\end{abstract}

\section{Introduction}
\label{sec:intro}
In recent years, dialogue systems have achieved fruitful results with neural conversation models in both open-domain generation~\cite{ritter2011data,sordoni2015neural,li2016deep,li2017adversarial,xu2017neural,zhang2018generating} and task-oriented completion~\cite{wen2015semantically,wen2017network,williams2017hybrid,bordes2016learning,su2018discriminative}. These methods empower lots of real-world dialogue applications such as Google Home and Apple Siri.

However, the utterance generation from dialogue systems still faces some critical challenges, including utterance blandness and incoherence~\cite{gao2018neural}. They are mainly caused by the objective function of the dialogue systems that prefer utterances with unconditionally high probability~\cite{li2016diversity}. 
We argue that in a meaningful and coherent dialogue, the change of utterance order will lead to a low-quality dialogue. However, most existing neural-based dialogue systems either encode the full dialogue history~\cite{li2017adversarial,xu2017neural} or only the current utterance~\cite{liu2018adversarial}. None of them explicitly models the sequential order and studies its criticality to the dialogue learning problem.

In this paper, we explore the sequential order within the dialogue as the self-supervised signal to guide meaningful and coherent dialogue learning. We introduce a self-supervised learning task, inconsistent order detection, to explicitly capture the order signal of the dialogue. The task is defined as, given a target utterance pair triple, the model is required to predict whether the triple is correctly ordered or not.
For instance, the utterance pair triple $\langle(Q_1,A_1), (Q_4,A_4), (Q_2, A_2)\rangle$ is misordered.
The key to solving this task is to model the utterance order based on the dialogue context effectively. 
But when directly encoding the full dialogue history along the temporal order, the model actually only focuses on the ending utterances, and earlier information is largely discarded~\cite{li2017adversarial}.
Thus, we propose a sampling-based \textbf{self-supervised network} ($\mathcal{SSN}$) to account for the forgetfulness problem and solve the inconsistent order detection task. In order to accurately predict if a target utterance triple is ordered or not, we randomly sample utterance triples from the dialogue history as the reference to incorporate the dialogue context. 
Since for the same target utterance triple, the sampled triple references are different at different iterations during training. It essentially approximates the full dialogue history without suffering from the forgetfulness issue.

To further utilize $\mathcal{SSN}$ in real dialogue learning, we propose to jointly learn $\mathcal{SSN}$ and the dialogue model via alternative training, where the output probability of $\mathcal{SSN}$ is treated as the order signal to evaluate the generated utterance. Moreover, the proposed approach can be applied to both open-domain and task-oriented dialogue learning, which indicates that $\mathcal{SSN}$ is a general and scalable approach for dialogue learning. Empirical results on two widely-used benchmark datasets, OpenSubtitles and Movie-Ticket Booking, show that our self-supervised network consistently improves the state-of-the-art (SOTA) neural-based dialogue training methods. In summary, our main contributions are three-fold:
\begin{itemize}
\item We introduce the task of inconsistent order detection, and propose a self-supervised learning network $\mathcal{SSN}$ to solve this task and explicitly model the crucial order information in dialogue.
\item We propose a general framework to jointly learn $\mathcal{SSN}$ and the dialogue models, where the sequential order in dialogues can be explicitly used to guide the utterance generation.
\item Our method advances the existing state-of-the-art dialogue systems in both open-domain and task-oriented scenarios.
\end{itemize}

\section{Related Work}
\label{sec:related}
\paragraph{Dialogue Learning}
Dialogue systems can be roughly classified into open-domain and task-oriented scenarios. In recent years, neural-based conversation models have shown great power in building dialogue systems~\cite{ritter2011data,sordoni2015neural,vinyals2015neural,serban2016building,luan2016lstm}. However, the utterances generated by neural-based dialogue systems still suffer from blandness and incoherence~\cite{gao2018neural}. To address these problems, \newcite{li2016diversity} propose a mutual information objective to infer the utterance generation. \newcite{serban2017hierarchical} and \newcite{zhang2018learning} further apply the latent variable models to generate more specific responses. Similar to some language generation tasks~\cite{lamb2016professor,yu2017seqgan}, 
Generative adversarial networks (GAN)~\cite{goodfellow2014generative} have also been adapted to learn a better objective function for the dialogue~\cite{li2017adversarial,xu2017neural,liu2018adversarial,su2018discriminative}. The discriminator in GAN is often used to evaluate the generated utterances and guide dialogue learning. However, these methods mainly focus on the surface information of generated utterances to guide the dialogue learning, and fail to consider the utterance connection within the dialogue history. In this paper, we focus on the sequential information of the dialogue and show that the unique sequential order in a meaningful and coherent dialogue contains more useful semantic information for dialogue learning.

\paragraph{Self-Supervised Learning}
Self-supervised learning, which aims to train a network on an auxiliary task where ground-truth is obtained automatically, has been successfully applied in computer vision. Many self-supervised tasks have been introduced to use non-visual but intrinsically correlated features to guide the visual feature learning~\cite{doersch2015unsupervised,wang2015unsupervised,pathak2016context}. As for natural language processing, predicting nearby words~\cite{mikolov2013distributed,mikolov2013efficient} is a self-supervised task to learn word embeddings. The language modeling is another line of self-supervision where a language model learns to predict the next word given the previous sequence~\cite{bengio2003neural,dai2015semi,peters2018deep}. Recently, \newcite{devlin2018bert} further proposes two self-supervised tasks, the masked language model and next sentence prediction, to learn sentence embeddings. \newcite{lample2019cross,liu2019multi} further extend these two tasks into multi-lingual and multi-task paradigms. ~\newcite{wang2019self} consider them at the sentence-level for extractive summarization. Our work is the first to consider the sequential order as the self-supervised signal in dialogue and we propose the self-supervised task of inconsistent order detection towards more coherent and relevant dialogue learning.

\begin{figure*}[t]
\centering
\includegraphics[width=0.8\columnwidth]{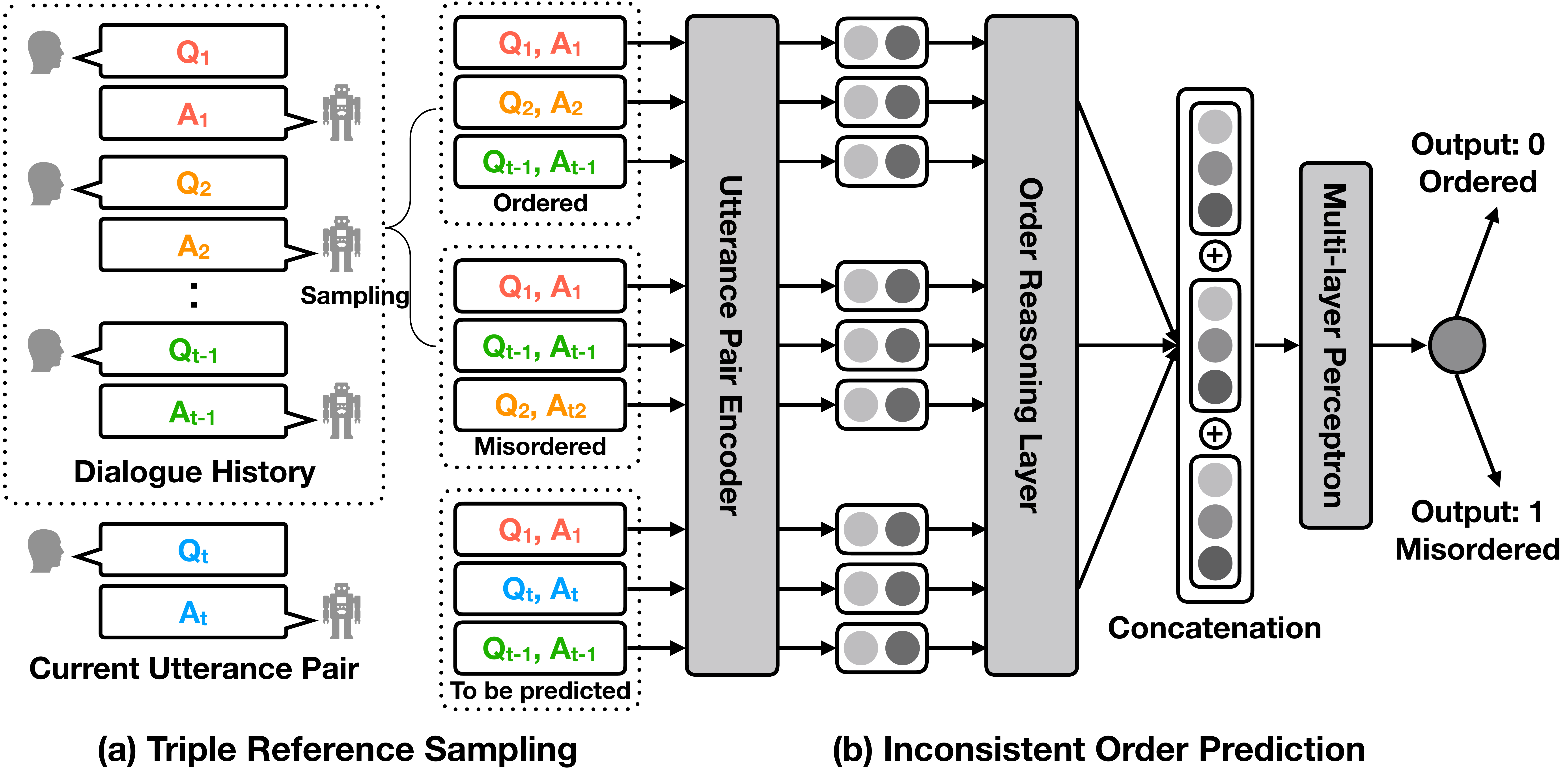}
\caption{The overview of our self-supervised network ($\mathcal{SSN}$) for inconsistent order detection. Given a target triple containing the current utterance pair $(Q_t, A_t)$ to be predicted, (a) we first sample triple references from previous dialogue history 
$\{(Q_1, A_1),\cdots, (Q_{t-1}, A_{t-1})\}$ in each iteration. The references can be ordered or misordered. (b) For each triple, it is transformed into the triple embedding. The concatenation of triple embeddings is fed into a MLP, and gives the probability based on the current sampling.} 
\label{fig:overview}
\end{figure*}

\section{Methods}
\label{sec:method}
In this section, we systematically describe how to utilize the internal sequential order of utterances as self-supervision for dialogue learning. In Section~\ref{subsec:method:oid}, we first introduce the task of inconsistent order detection, where the model needs to predict whether one sampled triple of the dialogue is correctly ordered or not. We then present an effective sampling-based approach, self-supervised network ($\mathcal{SSN}$), to learn to capture the important order signal and solve this task (see Section~\ref{subsec:method:ssn}). In the end, we show in Section~\ref{sec:method:dialogue} how $\mathcal{SSN}$ can contribute to both open-domain and task-oriented dialogue learning by modeling the inconsistent order detection. 

\subsection{Inconsistent Order Detection}
\label{subsec:method:oid}
The dialogue systems aim at conversing with the human in a meaningful and coherent way~\cite{gao2018neural}. Thus, the sequential order in dialogue data is an important signal for building a good dialogue system. Existing neural-based dialogue systems only consider this signal in a weak and implicit way, where they use hierarchical encoders to model the dialogue history~\cite{sordoni2015hierarchical,serban2016building,li2017adversarial,serban2017hierarchical,xing2018hierarchical}. However, we argue that these methods are mainly designed to model the overall semantic context information of the dialogue history but not good at modeling intermediate sequential order. Especially, the order signal is becoming weak as the number of dialogue turns increases. Thus, we propose the task of inconsistent order detection to force building models to capture this signal as self-supervision explicitly. Given a dialogue till the turn $t$, we can formulate it as $\{(Q_1, A_1), (Q_2, A_2), \cdots, (Q_t, A_t)\}$, where $(Q_t, A_t)$ is a pair of human-machine utterances. Then we can sample multiple triples of this dialogue as utterance pair triples using the following strategies:
\begin{itemize}
\item \textbf{Ordered triple sampling}: We sample a triple following the dialogue sequential order as $\langle(Q_i, A_i), (Q_j, A_j), (Q_k, A_k)\rangle$, where $i<j<k\leq t$.
\item \textbf{Misordered triple sampling}: The three utterance pairs are sampled in a triple as $\langle (Q_i, A_i), (Q_k, A_k), (Q_j, A_j)\rangle$, where $i<j<k\leq t$.
\end{itemize}
Note that when the current dialogue length $t<=2$, it is not enough to get a rational sampling for utterance pair triples. Thus, we add three extra shared padding utterance pairs $(Q_{-2}, A_{-2})$, $(Q_{-1}, A_{-1})$ and $(Q_0, A_0)$ ahead of all the dialogue data before sampling\footnote{Specifically, e.g., for the added padding utterance $Q_{-2}$, it is represented as a sequence of one same padding word $\{w^{(Q_{-2})}_1, w^{(Q_{-2})}_2, \cdots, w^{(Q_{-2})}_N\}$, where $N$ is the rounded-up averaged length of utterances in the dataset.}.

Based on above triple sampling strategies, we define the task of inconsistent order detection as: \emph{given a dialogue history $\{(Q_1, A_1), (Q_2, A_2), \cdots, (Q_t, A_t)\}$ and the target utterance pair $(Q_t, A_t)$ for evaluation, the model needs to predict whether the sampled triple $T$ containing $(Q_t, A_t)$ is ordered or not}. For instance, $\langle(Q_1, A_1), (Q_2, A_2), (Q_t, A_t)\rangle$ is ordered (output: 0), while $\langle(Q_1, A_1), (Q_t, A_t), (Q_2, A_2)\rangle$ is misordered (output: 1).

\subsection{Self-Supervised Network $\mathcal{SSN}$} 
\label{subsec:method:ssn}
We plan to build the model to solve the inconsistent order detection task, and explicitly capture the sequential order in dialogue. The overview of our approach is shown in Figure~\ref{fig:overview}. At each dialogue turn $t$, given a target triple containing the current utterance pair, we first sample triple references from the previous dialogue history to capture more semantic context in dialogue. The target triple and triple references are then transformed into embeddings using an utterance pair encoder and an order reasoning layer. Finally, the concatenation of embeddings is used for the final prediction. We then describe the $\mathcal{SSN}$ in detail as follows.

\subsubsection{Triple Reference Sampling}
\label{subsubsec:ref}
Given the task definition in Section~\ref{subsec:method:oid}, the model needs to predict whether there is inconsistent order in the target triple containing the current utterance pair $(Q_t, A_t)$. It is intuitive that if we can get more previous dialogue history, we may make a better prediction for inconsistent order. One trivial way is to encode the full previous dialogue history using a hierarchical network and make the prediction. However, \newcite{li2017adversarial} suggests that this structure actually focuses more on the final two preceding utterances instead of the whole history. The sequential order signal is very weak in this condition. We also report some similar results in Section~\ref{subsec:exp:intrin}.

Therefore, we propose a sampling-based approach to model the utterance order based on the dialogue context effectively. For each sampling operation, we sample two triple references $T'$ and $T''$ from the previous dialogue history $\{(Q_1, A_1), (Q_2, A_2), \cdots, (Q_{t-1}, A_{t-1})\}$ following the sampling strategies in Section~\ref{subsec:method:oid}. In general, we explore the following three combinations of reference sampling strategies for $T'$ and $T''$:
\begin{itemize}
\item $T'$ and $T''$ are sampled ordered references.
\item $T'$ and $T''$ are sampled misordered ones.
\item $T'$ is ordered while $T''$ is misordered.
\end{itemize}
Note that in our experiments, we choose one certain combination and keep using it for sampling the triple references for all the target triples.

\subsubsection{Objective Function}
Given the target triple embedding $T$ and the triple reference embedding $T'$ and $T''$, we use $\mathcal{SSN}$ to calculate the probability $p(T|T', T'') = \mathcal{SSN}(T, T', T'')$. We use the Binary Cross Entropy loss to train the model:
\begin{equation}
L = -\mathbb{E}(y\log p(T|T', T'')),
\end{equation}
where $y$ is the ground-truth label. 

Considering that for the same target triple $T$, the triple references are sampled $m$ times to approximate the full dialogue history. Then we can rewrite the loss function as
\begin{equation}
\label{equ:ssn-objective}
L = -\mathbb{E}(\frac{1}{m}\sum_{i=1}^m y\log (p^{(i)}(T|T^{(i)'}, T^{(i)''}))),
\end{equation}
where $T^{(i)'}, T^{(i)''}$ are the triple references of $i$-th sampling. This is essentially a Monte Carlo estimation and the model would effectively incorporate the dialogue context and capture the order information, avoiding from directly encoding the full dialogue history and the forgetfulness issue.

\subsubsection{Network Structure}
In this section, we demonstrate how $\mathcal{SSN}$ embeds both the target triple $T$ and triple reference $T'$ and $T''$ to generate $p(T|T', T'')$ in each sampling.

\paragraph{Utterance Pair Encoder}
First, given a utterance pair $(Q_t, A_t)$, we concatenate the $Q_t$ and $A_t$ as one sequence. The sequence is then fed into a bidirectional long short-term memory network (LSTM)~\cite{hochreiter1997long}, and the utterance pair embedding $\mathbf{U}_t$ is the concatenation of the final two states of the bi-LSTM:
\begin{equation}
\mathbf{U}_t =  \left[
 \begin{matrix}
   \overleftarrow{\mathbf{h}_1}\\
   \overrightarrow{\mathbf{h}_{N_t}} 
  \end{matrix}
  \right],
\end{equation}
where $N_t$ is the length of the concatenated utterance sequence.

\paragraph{Order Reasoning Layer}
After obtaining the utterance pair embeddings $(\mathbf{U}_i, \mathbf{U}_j, \mathbf{U}_k)$ of a sampled triple $T = <(Q_i, A_i), (Q_j, A_j), (Q_k, A_k)>$, we need to reason and predict whether there is inconsistent order or not. To simplify our model, we use a 3-step reasoning bi-LSTM with the max-pooling layer to perform the order reasoning:
\begin{equation}
\mathbf{T} =  \left[
 \begin{matrix}
   \text{max-pooling}(\overleftarrow{\mathbf{h}_1},\overleftarrow{\mathbf{h}_2},\overleftarrow{\mathbf{h}_3})\\
   \text{max-pooling}(\overrightarrow{\mathbf{h}_1},\overrightarrow{\mathbf{h}_2},\overrightarrow{\mathbf{h}_3})
  \end{matrix}
  \right],
\end{equation}
where the input of each time step in bi-LSTM is one utterance pairs embedding, and $\mathbf{T}$ is the final embedding of the given triple.

Given the target triple embedding $\mathbf{T}$ and the triple reference embedding $\mathbf{T}'$ and $\mathbf{T}''$, the concatenation of these three embeddings is fed into a multi-layer perceptron, returning the probability $p(T|T', T'')$ of the triple is ordered (approaching 0) or misordered (approaching 1).

\subsection{Self-Supervised Network for Dialogue}
\label{sec:method:dialogue}
In this section, we explain how the $\mathcal{SSN}$ can be applied to the current dialogue system in both open-domain and task-oriented scenarios.

Suppose we have a dialogue system the the history $\{(Q_1, A_1),\cdots,(Q_{t-1},A_{t-1})\}$, at turn $t$, the system generate the utterance $A_t$ based on the $Q_t$. We can sample a misordered target triple $T$ containing $(Q_t, A_t)$. Following the assumption that the sequential order in a meaningful and coherent dialogue should be unique, the $\mathcal{SSN}$ will be easy to detect the inconsistent order in $T$ if the generated $A_t$ is good. Otherwise, the $A_t$ may be of low quality. Therefore, we take a two-step sampling approach to evaluate the generated utterance $A_t$ using $\mathcal{SSN}$. First, a misordered target triple $T$ containing $(Q_t, A_t)$ is sampled. Then we further sample triple references $T'$ and $T''$ as in Section~\ref{subsubsec:ref} and how easily the misorder in the sampled $T$ can be detected is measured as $\mathbb{E}_{T',T''}(p(T|T',T'')$. Based on the generated utterance $A_t$, we can sample multiple misordered $T$, and we set the following expectation to measure the probability that $A_t$ is a good generated utterance:

\begin{equation}
p^*_{\mathcal{SSN}} = \mathbb{E}_{\text{misordered}\;T}\mathbb{E}_{T',T''}(p(T|T',T'')).
\end{equation}

\begin{figure}[t]
\centering
\includegraphics[width=0.85\columnwidth]{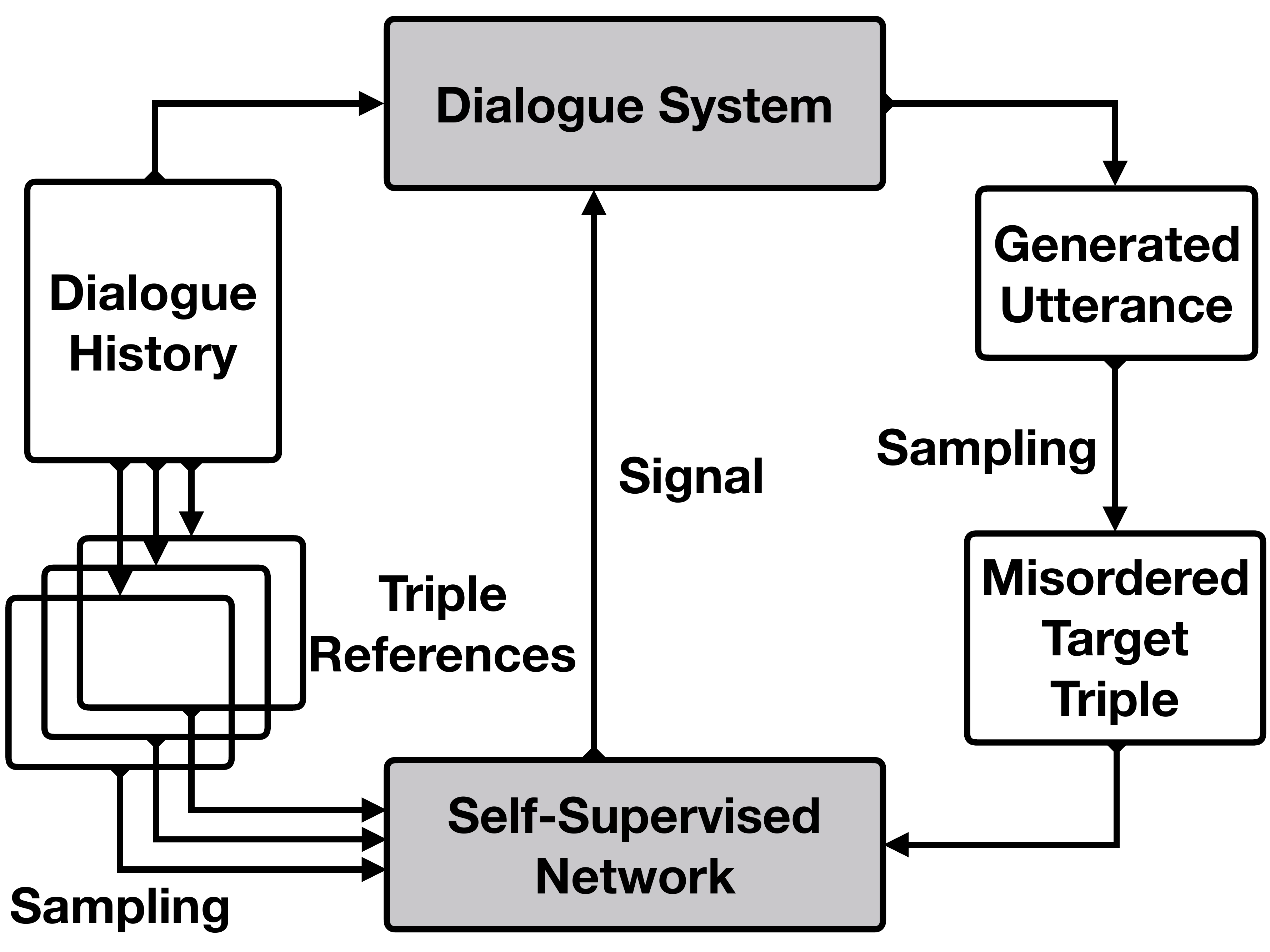}
\caption{The general framework for dialogue learning with self-supervised network.\label{fig:optim}} 
\end{figure}

In this way, we can view human-generated utterances as good ones, and machine-generated utterances as bad ones. Then we can use the adversarial training methods~\cite{goodfellow2014generative,li2017adversarial,xu2017neural,su2018discriminative} to train the dialogue system, where $\mathcal{SSN}$ can give clear order-based signal to guide the generator $G$ in the system. The framework of using $\mathcal{SSN}$ with the two-step sampling in real dialogue systems are shown in Figure~\ref{fig:optim}. The objective function then can be formulated as:
\begin{equation}
\label{equ:minmax}
\begin{aligned}
\min_{\theta_G}\max_{\theta_{\mathcal{SSN}}}&\mathbb{E}_{real}[\log p^*_{\mathcal{SSN}}(x)] \\ 
& + \mathbb{E}_{gen}[\log(1-p^*_{\mathcal{SSN}}(G(.)))],
\end{aligned}
\end{equation}
where $\theta_G$ and $\theta_{\mathcal{SSN}}$ are the parameters of the generator $G$ and $SSN$ in the dialogue systems separately. The $x$ stands for real human-generated utterances, which $G(.)$ represents machine-generated ones. The $G$ and $\mathcal{SSN}$ are alternately updated during training. We further describe the details in open-domain and task-oriented scenarios separately.

\subsubsection{Open-Domain Dialogue Learning}
The open-domain dialogue task is, given a dialogue history consisting of a sequence of dialogue utterances $\{(Q_1, A_1), \dots, (Q_{t-1}, A_{t-1})\}$, and current $Q_t$, the model needs to generate a response utterance $A_t$. We consider the adversarial training~\cite{li2017adversarial,xu2017neural} for dialogue generation systems. Following the previous approach~\cite{vinyals2015neural,serban2016building,luan2016lstm,li2017adversarial}, we use the \sts model for response generation as the generator $G$. The \sts first transforms the dialogue history into an embedding using an encoder recurrent network. Conditioned on the history embedding, another decoder recurrent network then computes the probability of tokens at each generation step of the response using a softmax function.

As for the discriminator $D$, in previous methods, the discriminator directly takes the response utterance $A_t$ with or without the full dialogue history, and predicts whether it is human-generated (output: 1) or machine-generated (output: 0). The probability of being human-generated is set as the reward to update the $G$ using the REINFORCE algorithm~\cite{williams1992simple}. As for our $\mathcal{SSN}$, the reward $R$ is set as $R=p^*_{\mathcal{SSN}}$.

\subsubsection{Task-Oriented Dialogue Learning}
The task-oriented dialogue, usually formulated as a reinforcement learning problem, aims to build a dialogue agent to interact with real users and learn the policy to complete the slot-filling task~\cite{jurafsky2014speech}. While the real-user interaction is expensive and time-consuming, in this scenario, the dialogue systems are often trained with user simulators~\cite{schatzmann2006survey,li2016user}. However, due to the complexity of real conversations and biases in the design of user simulators, the quality of simulated utterances is unstable. \newcite{su2018discriminative} propose an adversarial learning approach to differentiate simulated experience from real experience. Following the similar assumption that real-user interactions should be meaningful and coherent, we implement our $\mathcal{SSN}$ instead of the conventional discriminator $D$ to select high-quality stimulated utterances in the task-oriented dialogue systems.

In this scenario, the generator $G$ is the world model which produces simulated user experience, and the $\mathcal{SSN}$ focuses on scoring the simulated user experience $Q_t$ during the training process. Thus, instead of sampling and encoding utterance pairs $(Q_t, A_t)$, here we only use the user utterance $Q_t$ in $\mathcal{SSN}$. We keep other parts of the $\mathcal{SSN}$ remain the same as in Section~\ref{subsec:method:ssn}. Because the world model $G$ is updated using the multi-task learning without the reward from the $\mathcal{SSN}$, the objective function of the $\mathcal{SSN}$ in Equation~\ref{equ:minmax} can be rewritten as the following during the mini-batch training:
\begin{equation}
\frac{1}{b}\sum^b_{i=1}[\log p^*_{\mathcal{SSN}}(x^{(i)}) + \log(1-p^*_{\mathcal{SSN}}(G(.)^{(i)}))],
\end{equation}
where $b$ represents the batch size.

\section{Experiments}
\label{sec:exp}

\subsection{Intrinsic Evaluation}
\label{subsec:exp:intrin}
Before we deploy the self-supervised network into real dialogue systems, we first test the model architectures for reliability. We randomly choose $40K$ balanced ordered and misordered utterance pair triples from the OpenSubtitles~\cite{tiedemann2009news} dataset, and train the $\mathcal{SSN}$ to solve this $2$-class classification. We sample another $1K$ balanced triples for testing. We also consider a baseline model, where the target triple is encoded by $\mathcal{SSN}$, and the previous dialogue history is encoded by a hierarchical LSTM. The concatenation of two embeddings is used for the final prediction. 
Because our $\mathcal{SSN}$ is a sampling-based approach, we report the average prediction accuracy of $5$ runs on the $2$-class classification as shown in Table~\ref{tab:intrinsic}.

\begin{table}[t]
\small
\begin{center}
\begin{tabular}{lc}
\toprule
\textbf{Reference Strategy of} $\mathcal{SSN}$ & \textbf{Average Accuracy}\\
\midrule
All history by hierarchical LSTM & $.694 \;(.006)$\\
\midrule
w/o Refers & $.670 \;(.011)$\\
2*Ordered Refers& $.740 \;(.031)$\\
2*misordered Refers& $.744 \;(.029)$\\
1*Ordered + 1*misordered Refers & $\mathbf{.856 \;(.017)}$\\
\bottomrule
\end{tabular}
\end{center}
\caption{\label{tab:intrinsic}The intrinsic evaluation results. The numbers in brackets stand for deviation. Refers: Reference Triples.} 
\end{table}

From the results, we can observe that: (1) The conventional hierarchical LSTM is not suitable for this task, and this baseline only shows a marginal improvement compared with the strategy that only considers target triple without any history. The results also match previous findings~\cite{li2017adversarial}, where they suggest that only the last two proceeding utterances in the hierarchical network are semantically significant. (2) As for our $\mathcal{SSN}$, it is safe to tell that reference triples can be a tremendous supplement to the inconsistent order detection. It is not surprising because by adding reference triples, the $\mathcal{SSN}$ will know more information of semantic context within the dialogue. Especially when having both ordered and misordered references, the $\mathcal{SSN}$ has the highest classification accuracy. This also shows that the sampling strategy, 1*Ordered + 1*misordered references, is the most reliable structure for real dialogue systems. Thus, for the rest of the experiments, we directly use the $\mathcal{SSN}$ with one ordered and one misordered references strategy to achieve the best performance.

\subsection{Open-Domain Dialogue Learning}
\label{subsec:exp:open}
\paragraph{Dataset}
Following the previous studies~\cite{vinyals2015neural,li2017adversarial,xu2017neural}, we choose the widely-used OpenSubtitles~\cite{tiedemann2009news} dataset to evaluate different methods. The OpenSubtitles dataset contains movie scripts organized by characters, where we follow~\newcite{li2016deep} to retain subtitles containing 5-50 words.

\paragraph{Baselines} We consider the following two popular adversarial methods for dialogue learning as the baselines:
\begin{itemize}
\item \textbf{REGS}~\cite{li2017adversarial}: The discriminator $D$ takes the full dialogue history by a hierarchical LSTM, and the Monte Carlo search is implemented to obtain rewards for every generation step to update the generator $G$.
\item \textbf{AEL}~\cite{xu2017neural}: The discriminator $D$ only encodes the currently generated utterance by a CNN model and the generator $G$ is optimized using an approximate embedding layer.
\end{itemize}

\begin{table}[t]
\small
\begin{center}
\begin{tabular}{lccc}
\toprule
\textbf{Separated $G$/$D$} & $D$-REGS & $D$-AEL & $D$-$\mathcal{SSN}$\\
\midrule
$G$-REGS & $.094$ & $.087$ & $.041$ \\
$G$-AEL & $.146$ & $.128$ & $.093$ \\
$G$-$\mathcal{SSN}$ & $\mathbf{.203}$ & $\mathbf{.185}$ & $\mathbf{.162}$ \\
\bottomrule
\end{tabular}
\end{center}
\caption{\label{tab:cross}The cross evaluation of adversarial success rate on different generators and discriminators. Please refer to Section~\ref{open:auto} Adversarial Evaluation for explanations.}
\end{table}

\begin{table}[t]
\small
\begin{center}
\begin{tabular}{lcc}
\toprule
\textbf{Model} & \textbf{distinct-1} & \textbf{distinct-2}\\
\midrule
REGS & $0.0217$ & $0.0695$ \\
AEL & $0.0311$ & $0.0948$ \\
$\mathcal{SSN}$ & $\mathbf{0.0393}$ & $\mathbf{0.1126}$ \\
\bottomrule
\end{tabular}
\end{center}
\caption{\label{tab:distinct}The automatic evaluation of generated utterances on distinct-1 and distinct-2 metrics. Please refer to Section~\ref{open:auto} Automatic Evaluation for explanations.}
\end{table}

\begin{table}[t]
\small
\begin{center}
\begin{tabular}{lccc}
\toprule
\textbf{Win} & REGS & AEL & $\mathcal{SSN}$\\
\midrule
Single-turn Percentage & $.095$ & $.192$ & $\mathbf{.713}$\\
Multi-turn Percentage & $.025$ & $.171$ & $\mathbf{.804}$\\
\bottomrule
\end{tabular}
\end{center}
\caption{\label{tab:huamn}The human evaluation of generated utterances in three methods. The result here is statistically significant with $p < 0.01$ according to sign test. Please refer to Section~\ref{open:auto} Human Evaluation for explanations.}
\end{table}

\begin{table*}[ht]
\small
\setlength{\tabcolsep}{5pt}
\begin{center}
\begin{tabular}{lcccccccccc}
\toprule
\multirow{2}{*}{\textbf{Agent}} & \multirow{2}{*}{\textbf{\tabincell{c}{Planning \\ Steps}}} & \multicolumn{3}{@{}c}{\textbf{Epoch 100}} & \multicolumn{3}{@{}c}{\textbf{Epoch 200}} & \multicolumn{3}{@{}c@{}}{\textbf{Epoch 300}}\\
\cmidrule(lr){3-5}\cmidrule(lr){6-8}\cmidrule(lr){9-11}
 & & \textbf{Succ} & \textbf{Reward} & \textbf{Turns} & \textbf{Succ} & \textbf{Reward} & \textbf{Turns} & \textbf{Succ} & \textbf{Reward} & \textbf{Turns} \\
\midrule
D3Q & \multirow{4}{*}{5} & .7467 & 43.59 & 14.03 & .6800 & 34.64 & 15.92 & .7200 & 40.85 & 13.11 \\
D3Q-$\mathcal{SSN}$ & & \textbf{.7600} & \textbf{45.71} & \textbf{13.52} & \textbf{.7400} & \textbf{42.93} & \textbf{14.80} & \textbf{.7633} & \textbf{46.16} & 15.24 \\
D3Q (fixed $\theta_D$) & & .6800 & 33.86 & 17.48 & .7000 & 36.57 & 16.85 & .6933 & 35.67 & 17.06 \\
D3Q-$\mathcal{SSN}$ (fixed $\theta_{\mathcal{SSN}}$) & & .6633 & 32.04 & 16.21 & .7133 & 36.71 & 17.74 & .7067 & 36.03 & \textbf{12.91} \\
\midrule
D3Q & \multirow{4}{*}{10} & .6333 & 28.99 & 16.01 & .7000 & 37.24 & \textbf{15.52} & .6667 & 33.09 & 15.83 \\
D3Q-$\mathcal{SSN}$ & & \textbf{.7800} & \textbf{48.71} & 15.84 & \textbf{.8733} & \textbf{56.15} & 19.57 & \textbf{.8067} & \textbf{50.29} & 16.48 \\
D3Q (fixed $\theta_D$) & & .7133 & 36.36 & 20.48 & .8400 & 54.87 & 20.48 & .7400 & 42.89 & 13.81 \\
D3Q-$\mathcal{SSN}$ (fixed $\theta_{\mathcal{SSN}}$) & & .7367 & 42.30 & \textbf{14.79} & .8300 & 52.92 & 18.16 & .7933 & 48.05 & \textbf{13.73} \\
\bottomrule
\end{tabular}
\end{center}
\caption{\label{tab:d3q}The experimental results of different dialogue agents at training epoch = $\{100, 200, 300\}$. Each number is averaged over 3 runs, and each run tested on 50 dialogues. The D3Q-$\mathcal{SSN}$ denotes the D3Q agent where our proposed $\mathcal{SSN}$ replaces the discriminator. The ``fixed $\theta_D/\theta_{\mathcal{SSN}}$" indicates the discriminator/$\mathcal{SSN}$ is pre-trained and fixed during the training process. Succ: Success Rate. Reward: Average Reward. Turns: Average Turns.}
\end{table*}

\paragraph{Implementation Details}
We follow the most of parameters in~\newcite{li2017adversarial,xu2017neural} to make a fair comparison. For the generator model $G$, we adopt the same \sts model~\cite{sutskever2014sequence} with an attention mechanism~\cite{bahdanau2014neural,luong2015effective} for our approach and baselines. We approximate the dialogue history for $G$ using the concatenation of two preceding utterances following the~\newcite{li2017adversarial}. To train the generator $G$, we use the REINFORCE algorithm~\cite{williams1992simple} to maximize the expected reward of generated utterances. We also implement the Monte Carlo search to give rewards for each generation step. To accelerate the sampling process, we use multiple GPUs to parallelize and distribute the jobs. As for the $\mathcal{SSN}$, it first gets pre-trained using sampled data from OpenSubtitiles, and then iteratively updated during the min-max adversarial training process. The dimension of the utterance embeddings is $128$. The hidden size is $256$ for utterance encoding bi-LSTM and $1024$ for triple reasoning bi-LSTM. The MLP has a single hidden layer of size $512$.

\paragraph{Adversarial Evaluation}
Here we use adversarial success rate (AdverSuc), which is the fraction of instances where a $G$ is capable of fooling the $D$, to evaluate different methods. Higher values of AdverSuc for a dialogue system usually lead to a better response generator. After training three $(G,D)$ using REGS, AEL and $\mathcal{SSN}$, we sample $4K$ dialogue history and use three trained generators to generate response utterances. These machine-generated utterances are then fed into three trained discriminators to see if they are indistinguishable from human-generated ones. The cross evaluation of AdverSuc is shown in Table~\ref{tab:cross}.

From the results, we can observe that: (1) Our trained generator achieve higher AdverSuc in three discriminators, which shows that the generator in our approach can generate more human-like utterance responses. (2) The generators of the other two methods have a noticeable drop in AdverSuc when evaluating on our $\mathcal{SSN}$-based discriminator. This demonstrates that our self-supervised policy for discriminating utterances is successful. (3) The REGS method with full dialogue history encoded performs worse than the AEL that only considers the current utterances. We think this indicates that without explicitly stating the guiding signal, both the generator and the discriminator can be lost about figuring out a good objective function during the training process even when encoding the full history.

\paragraph{Automatic Evaluation}
\label{open:auto}
For automatic evaluations, we use the two commonly accepted metrics distinct-1 and distinct-2. The distinct-1 and distinct-2, proposed by~\newcite{li2016diversity}, are two ways to measure the degree of diversity by calculating the number of distinct unigrams and bigrams in the generated response utterances. The evaluation results are reported in Table~\ref{tab:distinct}. The results show that based on the distinct-1 and distinct-2 metrics, the generator trained in our approach can generate relatively more diverse responses. The results are attractive considering that we do not explicitly use a diversity-guided objective function during the training process. We think the reason is that the diverse utterances are easier to reserve the order information. In previous methods, the discriminator $D$ only gives good or bad signals to response generator $G$, and the $G$ has to figure out what is an acceptable response by itself. As for our $\mathcal{SSN}$, it explicitly forces the $G$ to generate responses that will have unique orders in dialogue, which leads to more diverse utterances.

\paragraph{Human Evaluation}
For human evaluation, we follow protocols in~\newcite{li2016diversity} and employing crowd-sourced judges from the Amazon Mechanical Turk to evaluate a random sample of 1000 unique generated utterances from three generators in the OpenSubtitles test dataset. We present both the input dialogue history and the generated responses to 5 judges and ask them to decide which one of the three results is the be.ts Ties are not permitted. We consider both single-turn and multi-turn for the evaluation. The results are shown in Table~\ref{tab:huamn}. Evidently, the generator trained in our method shows a significant improvement in the quality of generated sentences. The gain is even higher in the multi-turn setting than the single-turn setting. This is because when only considering the single-turn dialogue, the information encoded in three methods will be similar.

\subsection{Task-Oriented Dialogue Learning}
\label{subsec:exp:task}

\paragraph{Dataset} Following the previous work~\cite{peng2018deep,su2018discriminative}, we use the same Movie-Ticket Booking dataset collected from Amazon Mechanical Turk for evaluation. The dataset is manually labeled based on a schema defined by domain experts consisting of $11$ intents and $16$ slots in the full domain setting. In total, the dataset has $280$ annotated dialogues with an average length of approximately 11 turns. In this scenario, the goal of dialogue systems is to help the user complete the tasks through the conversation.

\paragraph{Baselines} We compare our $\mathcal{SSN}$-based discriminator within the state-of-the-art task-oriented dialogue policy learning approach, Discriminative Deep Dyna-Q (D3Q)~\cite{su2018discriminative}. At each turn, the D3Q agent takes $S$ planning steps interacting with the simulator and store stimulated user experiences based on the scoring of the discriminator. The stimulated user experiences are generated by the world model, which can be viewed as the generator $G$ in our case. We replace the conventional discriminator $D$ of D3Q with our $\mathcal{SSN}$.

\paragraph{Implementation Details} For a fair comparison, we remain most of the parameters in the D3Q algorithm the same as in~\newcite{su2018discriminative}. In the self-supervised network, the dimension of the utterance embeddings is $80$. The hidden size is $128$ for utterance encoding bi-LSTM and $512$ for triple reasoning bi-LSTM. The MLP has a single hidden layer of size $128$. We use the simulator\footnote{\url{https://github.com/MiuLab/TC-Bot}} as in~\newcite{li2016user} to generate user utterances, and the threshold interval is set to a range between $0.45$ and $0.55$.

\paragraph{Results} The experimental results of different agents at training epoch are shown in Table~\ref{tab:d3q}. From the results, we can observe that: (1) The D3Q-$\mathcal{SSN}$  outperform the D3Q in the most of cases, which shows that our $\mathcal{SSN}$-based discriminator can improve the ability to recognize the high-quality stimulated user experiences. (2) When the planning step increases in D3Q, the performance shows an apparent drop. This is because the discriminator $D$ in the original D3Q agent keeps lots of low-quality stimulated user experiences, which significantly degrade the performance of the D3Q agent. As for our $\mathcal{SSN}$, we can see some performance improvement even when using $10$-step planning. This substantially means that our $\mathcal{SSN}$ has a better ability to select the good simulated user experiences, especially in the multi-turn dialogue cases.

\section{Conclusion}
\label{sec:conclusion}
In this paper, we introduce a self-supervised task, inconsistent order detection, to explicitly capture the order signal of the dialogue. While previous methods suffer from forgetfulness problem when modeling dialogue history, we further propose a sampling-based self-supervised network $\mathcal{SSN}$, to approximately encoding the dialogue history and highlight the order signal. We also show how our $\mathcal{SSN}$ can contribute to real dialogue learning. Empirically, our method advances the previous state-of-the-art dialogue systems in both open-domain and task-oriented scenarios. Theoretically, we believe this self-supervision can be generalized to other types of temporal order in different NLP tasks.

\bibliography{acl2019}
\bibliographystyle{acl_natbib}
\end{document}